\documentclass{article}

\usepackage{bm}
\usepackage[utf8]{inputenc} 
\usepackage[T1]{fontenc}    
\usepackage{hyperref}       
\usepackage{url}            
\usepackage{booktabs}       
\usepackage{amsfonts}       
\usepackage{nicefrac}       
\usepackage{microtype}      
\usepackage{cancel}
\hypersetup{colorlinks=true}

\usepackage[ruled]{algorithm2e}
\usepackage{amsmath, amssymb, amsthm}
\usepackage{graphicx}
\usepackage{enumitem}
\usepackage{multirow}
\usepackage{stmaryrd}
\usepackage[dvipsnames]{xcolor}
\usepackage{scalerel}
\usepackage{subcaption}
\usepackage{wrapfig}
\usepackage{arydshln}

\graphicspath{{figures/}}
\setlist[itemize]{label={--},itemsep=0pt,topsep=0pt,partopsep=0pt}
\AtBeginDocument{%
  \setlength{\abovedisplayskip}{2pt}
  \setlength{\belowdisplayskip}{2pt}
}
\definecolor{mygreen}{RGB}{0, 153, 0}
\definecolor{myorange}{RGB}{204, 102, 0}
\definecolor{tsfgreen}{RGB}{0, 204, 0}
\definecolor{tsfblue}{RGB}{0, 102, 204}
\definecolor{darkpage}{RGB}{33, 33, 33}
\definecolor{lighttext}{RGB}{239, 239, 239}
\definecolor{themecyan}{RGB}{76, 157, 150}

\newcommand{\red}[1]{\textcolor{red}{#1}}
\newcommand{\green}[1]{\textcolor{mygreen}{#1}}
\newcommand{\blue}[1]{\textcolor{blue}{#1}}
\newcommand{\orange}[1]{\textcolor{myorange}{#1}}
\newcommand{\tsfblue}[1]{\textcolor{tsfblue}{#1}}
\newcommand{\tsfgreen}[1]{\textcolor{tsfgreen}{#1}}

\DeclareMathOperator*{\softmax}{softmax}

\DeclareMathOperator{\id}{Id}
\DeclareMathOperator{\diag}{diag}

\def\RR{{\mathbb R}}

\usepackage[normalem]{ulem}

    \def\cT{{\mathcal T}}

  \def\cL{{\mathcal L}}

\SetKwInput{KwInit}{Initialization}

\usepackage[nonatbib,final]{neurips_2020}

\title{Deep Transformation-Invariant Clustering}

\author{Tom Monnier\qquad\qquad Thibault Groueix\qquad\qquad Mathieu Aubry\\\\
  LIGM, Ecole des Ponts, Univ Gustave Eiffel, CNRS, France\\
  \texttt{\{tom.monnier,thibault.groueix,mathieu.aubry\}@enpc.fr} \\
}

\begin{document}

\maketitle

\begin{abstract}
  Recent advances in image clustering typically focus on learning better deep 
  representations. In contrast, we present an orthogonal approach that does not rely on 
  abstract features but instead learns to predict transformations and performs clustering 
  directly in pixel space. This learning process naturally fits in the gradient-based 
  training of K-means and Gaussian mixture model, without requiring any additional loss or 
  hyper-parameters. It leads us to two new deep transformation-invariant clustering 
  frameworks, which jointly learn prototypes and transformations. More specifically, we use 
  deep learning modules that enable us to resolve invariance to spatial, color and 
  morphological transformations. Our approach is conceptually simple and comes with several 
  advantages, including the possibility to easily adapt the desired invariance to the task 
  and a strong interpretability of both cluster centers and assignments to clusters.  We 
  demonstrate that our novel approach yields competitive and highly promising results on 
  standard image clustering benchmarks. Finally, we showcase its robustness and the 
  advantages of its improved interpretability by visualizing clustering results over real 
  photograph collections.
\end{abstract}

\section{Introduction}

Gathering collections of images on a topic of interest is getting easier every day: simple 
tools can aggregate data from social media, web search, or specialized websites and filter it 
using hashtags, GPS coordinates, or semantic labels. However, identifying visual trends in 
such image collections remains difficult and usually involves manually organizing images or 
designing an ad hoc algorithm.  Our goal in this paper is to design a clustering method which 
can be applied to such image collections, output a visual representation for each cluster and 
show how it relates to every associated image.

Directly comparing image pixels to decide if they belong to the same cluster leads to poor 
results because they are strongly impacted by factors irrelevant to clustering, such as exact 
viewpoint or lighting. Approaches to obtain clusters invariant to these transformations can 
be broadly classified into two groups. A first set of methods extracts invariant features and 
performs clustering in feature space.  The features can be manually designed, but most 
state-of-the-art methods learn them directly from data. This is challenging because images 
are high-dimensional and learning relevant invariances thus requires huge amounts of data.  
For this reason, while recent approaches perform well on simple datasets like MNIST, they 
still struggle with real images. Another limitation of these approaches is that learned 
features are hard to interpret and visualize, making clustering results difficult to analyze. 
A second set of approaches, following the seminal work of Frey and Jojic on 
transformation-invariant clustering~\cite{freyEstimatingMixtureModels1999,
freyFastLargescaleTransformationinvariant2002, 
freyTransformationinvariantClusteringUsing2003}, uses explicit transformation models to align 
images before comparing them. These approaches have several potential advantages: (i) they 
enable direct control of the invariances to consider; (ii) because they do not need to 
discover invariances, they are potentially less data-hungry; (iii) since images are 
explicitly aligned, clustering process and results can easily be visualized.  However, 
transformation-invariant approaches require solving a difficult joint optimization problem.  
In practice, they are thus often limited to small datasets and simple transformations, such 
as affine transformations, and to the best of our knowledge they have never been evaluated on 
large standard image clustering datasets.

In this paper, we propose a deep transformation-invariant (DTI) framework that enables to 
perform transformation-invariant clustering at scale and uses complex transformations.  Our 
main insight is to jointly learn deep alignment and clustering parameters with a single loss, 
relying on the gradient-based adaptations of 
K-means~\cite{macqueenMethodsClassificationAnalysis1967} and GMM 
optimization~\cite{dempsterMaximumLikelihoodIncomplete1977}.  Not only is predicting 
transformations more computationally efficient than optimizing them, but it enables us to use 
complex color, thin plate spline and morphological transformations without any specific 
regularization.  Because it is pixel-based, our deep transformation-invariant clustering is 
also easy to interpret: cluster centers and image alignments can be visualized to understand 
assignments. Despite its apparent simplicity, we demonstrate that our DTI clustering 
framework leads to results on par with the most recent feature learning approaches on 
standard benchmarks. We also show it is capable of discovering meaningful modes in real 
photograph collections, which we see as an important step to bridge the gap between 
theoretically well-grounded clustering approaches and semi-automatic tools relying on 
hand-designed features for exploring image collections, such as 
AverageExplorer~\cite{zhu2014averageExplorer} or ShadowDraw~\cite{lee2011shadowdraw}.

\begin{figure}
    \centering
    \begin{subfigure}[t]{\textwidth}
    \centering
    \qquad\qquad\qquad\,
    \includegraphics[width=0.8\columnwidth]{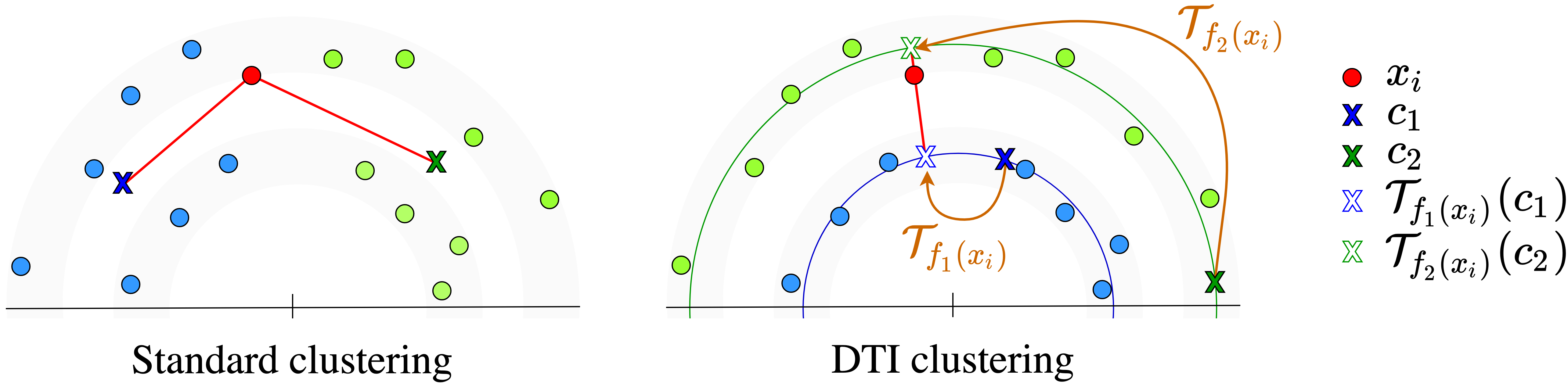}
    \caption{Classical versus Deep Transformation-Invariant clustering}
    \label{fig:method_a}
    \end{subfigure}
    \begin{subfigure}[t]{\textwidth}
    \centering
    \includegraphics[width=0.8\columnwidth]{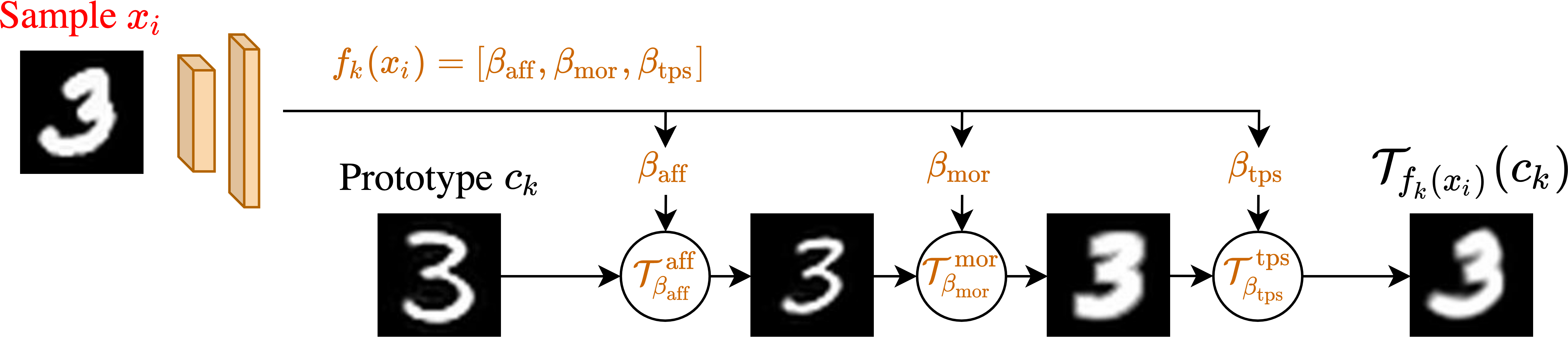}
    \caption{Deep transformation module $\mathcal{T}_{f_k}$}
    \label{fig:method_b}
    \end{subfigure}
    \begin{subfigure}[t]{\textwidth}
    \centering
    \vspace{0.4em}
    \includegraphics[width=0.49\columnwidth]{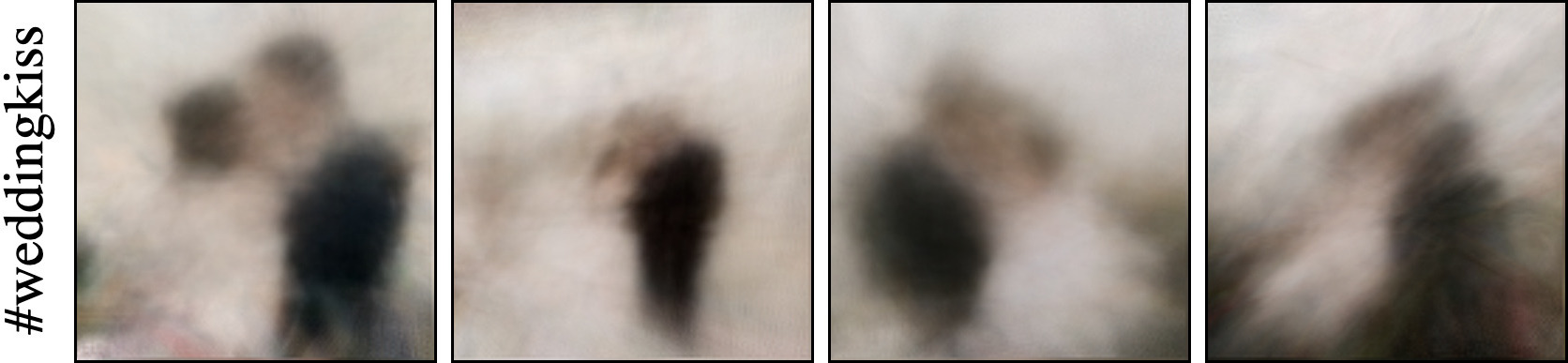}
    \includegraphics[width=0.49\columnwidth]{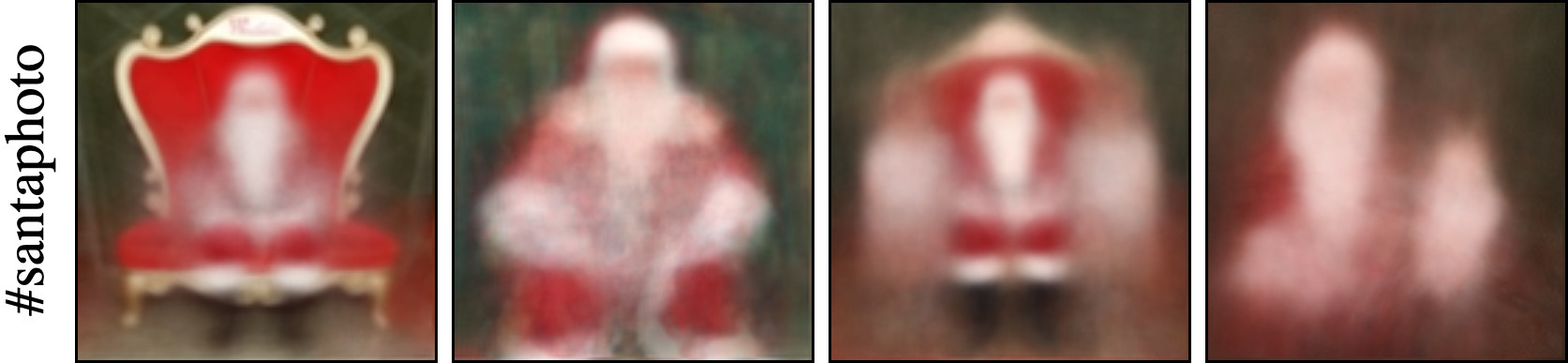}
    \caption{Prototypes learned from unfiltered Instagram images associated to different 
    hashtags}
    \label{fig:teaser}
    \end{subfigure}

    \caption{\textbf{Overview.}  \textbf{(a)} Given a sample \red{$x_i$} and prototypes 
      \blue{$c_1$} and \green{$c_2$}, standard clustering such as K-means assigns the sample 
      to the closest prototype.
    Our DTI clustering first aligns prototypes to the sample using a family of parametric 
    transformations - here rotations - then picks the prototype whose alignment yields the 
    smallest distance.
    \textbf{(b)} We predict alignment with deep learning.
    Given an image \red{$x_i$}, each parameter predictor~\orange{$f_k$} predicts parameters 
    for a sequence of transformations - here affine 
    \orange{$\mathcal{T}^{\,\textrm{aff}}_{\beta_{\textrm{aff}}}$}, morphological \orange{$ 
    \mathcal{T}^{\,\textrm{mor}}_{\beta_{\textrm{mor}}}$}, and thin plate spline 
    \orange{$\mathcal{T}^{\,\textrm{tps}}_{\beta_{\textrm{tps}}}$}
    - to align prototype {$c_k$} to \red{$x_i$}.
    \textbf{(c)} Examples of interpretable prototypes discovered from large images sets (15k 
  each) associated to hashtags in Instagram using our DTI clustering with 40 clusters.  Each 
cluster contains from 200 to 800 images.}
    \label{fig:overview_DTI}
    \vspace{-1.5em}
\end{figure}

We first briefly discuss related works in Section~\ref{sec:related}.  
Section~\ref{sec:approach} then presents our DTI framework (Fig.~\ref{fig:method_a}).  
Section~\ref{sec:learning} introduces our deep transformation modules and architecture 
(Fig.~\ref{fig:method_b}) and discuss training details. Finally, Section~\ref{sec:results} 
presents and analyzes our results (Fig.~\ref{fig:teaser}). 

\vspace{-0.7em}
\paragraph{Contributions.} In this paper we present:
\bgroup\setlength\parskip{0pt}
\begin{itemize}[itemsep=0pt,topsep=0pt,partopsep=0pt,parsep=0pt,leftmargin=5mm]
  \item  a deep transformation-invariant clustering approach that jointly learns to cluster 
    and align images,
  \item a deep image transformation module to learn spatial alignment,  color modifications 
    and for the first time morphological transformations,
  \item an experimental evaluation showing that our approach is competitive on standard image 
    clustering benchmarks, improving over state-of-the-art on Fashion-MNIST and SVHN, and 
    provides highly interpretable qualitative results even on challenging web image 
    collections.
\end{itemize}
\egroup
Code, data, models as well as more visual results are available on our
project \href{http://imagine.enpc.fr/~monniert/DTIClustering/}
{webpage}\footnote{http://imagine.enpc.fr/\textasciitilde monniert/DTIClustering/}.

\section{Related work}
\label{sec:related}

Most recent approaches to image clustering focus on learning deep image representations, or 
features, on which clustering can be performed. Common strategies include 
autoencoders~\cite{xieUnsupervisedDeepEmbedding2016, dizajiDeepClusteringJoint2017, 
jiangVariationalDeepEmbedding2017, kosiorekStackedCapsuleAutoencoders2019}, contrastive 
approaches~\cite{yangJointUnsupervisedLearning2016, changDeepAdaptiveImage2017,
shahamSpectralNetSpectralClustering2018}, 
GANs~\cite{chenInfoGANInterpretableRepresentation2016, zhouDeepAdversarialSubspace2018, 
mukherjeeClusterGANLatentSpace2019} and mutual information based 
strategies~\cite{huLearningDiscreteRepresentations2017, hausserAssociativeDeepClustering2018, 
jiInvariantInformationClustering2019}. Especially related to our work 
is~\cite{kosiorekStackedCapsuleAutoencoders2019} which leverages the idea of 
capsule~\cite{hintonTransformingAutoEncoders2011} to learn equivariant image features, in a 
similar fashion of equivariant models~\cite{lencUnderstandingImageRepresentations2015, 
taiEquivariantTransformerNetworks2019}. However, our method aims at being invariant to 
transformations but not at learning a representation.
 
\vspace{-0.3em}
Another type of approach is to align images in pixel space using a relevant family of 
transformations, such as translations, rotations, or affine transformations to obtain more 
meaningful pixel distances before clustering them. Frey and Jojic first introduced 
transformation-invariant clustering~\cite{freyEstimatingMixtureModels1999,
freyFastLargescaleTransformationinvariant2002, 
freyTransformationinvariantClusteringUsing2003} by integrating pixel permutations as a 
discrete latent variable within an Expectation Maximization 
(EM)~\cite{dempsterMaximumLikelihoodIncomplete1977} procedure for a mixture of Gaussians.  
Their approach was however limited to a finite set of discrete transformations.  {\it 
Congealing}\ generalized the idea to continuous parametric transformations, and in particular 
affine transformations, initially by using entropy 
minimization~\cite{miller2000learning,learned2005data}. A later version using least square 
costs~\cite{cox2008least,cox2009least} demonstrated the relation of this approach to the  
classical Lukas-Kanade image alignment algorithm~\cite{lucas1981iterative}.
In its classical version, congealing only enables to align all dataset images together, but 
the idea was extended to clustering~\cite{liu2009simultaneous, 
mattar2012unsupervised,li2016transformation}, for example using a Bayesian 
model~\cite{mattar2012unsupervised}, or in a spectral clustering 
framework~\cite{li2016transformation}. These works typically formulate difficult joint 
optimization problems and solve them by alternating between clustering and transformation 
optimization for each sample. They are thus limited to relatively small datasets and to the 
best of our knowledge were never compared to modern deep approaches on large benchmarks. Deep 
learning was recently used to scale the idea of congealing for global alignment of a single 
class of images~\cite{annunziata2019jointly} or time series~\cite{weber2019diffeomorphic}. 
Both works build on the idea of Spatial Transformer 
Networks~\cite{jaderbergSpatialTransformerNetworks2015} (STN) that spatial transformation are 
differentiable and can be learned by deep networks. We also build upon STN, but go beyond 
single-class alignment to jointly perform clustering. Additionally, we extend the idea to 
color and morphological transformations. We believe our work is the first to use deep 
learning to perform clustering in pixel space by explicitly aligning images. 

\section{Deep Transformation-Invariant clustering}\label{sec:approach}

In this section, we first discuss a generic formulation of our deep transformation-invariant 
clustering approach, then derive two algorithms based on 
K-means~\cite{macqueenMethodsClassificationAnalysis1967} and Gaussian mixture 
model~\cite{dempsterMaximumLikelihoodIncomplete1977}.

\vspace{-0.3em}
{\bf Notation:} In all the rest of the paper, we use the notation $a_{1:n}$ to refer to the 
set $\{a_1, \ldots, a_n\}$.

\subsection{DTI framework}\label{sec:dti}

Contrary to most recent image clustering methods which rely on feature learning, we propose 
to perform clustering in pixel space by making the clustering invariant to a family of 
transformations. We consider $N$ image samples $x_{1:N}$ and aim at grouping them in $K$ 
clusters using a \textit{prototype method}. More specifically, each cluster $k$ is defined by 
a prototype $c_k$, which can also be seen as an image, and prototypes are optimized to 
minimize a loss $\mathcal{L}$ which typically evaluates how well they represent the samples.  
We further assume that $\mathcal{L}$ can be written as a sum of a loss $l$ computed over each 
sample:%
\begin{equation}\label{eq:basic}
  \mathcal{L}(c_{1:K})=\sum_{i=1}^N l(x_i,\,\{c_{1}, \ldots,c_K\}).
\end{equation}
Once the problem is solved, each sample $x_i$ will be associated to the closest prototype.

Our key assumption is that in addition to the data, we have access to a group of parametric
transformations $\{\cT_\beta,\, \beta\in B \}$ to which we want to make the clustering 
invariant. For example, one can consider $\beta \in \mathbb{R}^6$ and $\cT_\beta$ the 2D 
affine transformation parametrized by $\beta$. Other transformations are discussed in 
Section~\ref{sec:transfo}. Instead of finding clusters by minimizing the loss of 
Equation~\ref{eq:basic}, one can minimize the following transformation-invariant loss:%
\begin{equation}
  \mathcal{L}_{\textrm{TI}}(c_{1:K})=\sum_{i=1}^N \min_{\beta_{1:K}}\  l(x_i,\,
  \{\cT_{\beta_1}(c_1), \ldots,\cT_{\beta_K}(c_K)\}).
    \label{eq:ti}
\end{equation}
In this equation, the minimum over $\beta_{1:K}$ is taken for each sample independently. This 
loss is invariant to transformations of the prototypes (see proof in 
Appendix~\ref{sec:proof}).  Also note there is not a single optimum since the loss is the 
same if any prototype $c_k$ is replaced by $\cT_{\beta}(c_k)$ for any $\beta\in B$. If 
necessary, for example for visualization purposes, this ambiguity can easily be resolved by 
adding a small regularization on the transformations. The optimization problem associated to 
$\mathcal{L}_{\textrm{TI}}$ is of course difficult.
A natural approach, which we use as baseline (noted TI), is to alternatively minimize over 
transformations and clustering parameters. We show that performing such optimization using a 
gradient descent can already lead to improved results over standard clustering but is 
computationally expensive. 

\vspace{-0.3em}
We experimentally show it is faster and actually better to instead learn $K$ (deep) 
predictors $f_{1:K}$ for each prototype, which aim at associating to each sample $x_i$ the 
transformation parameters $f_{1:K}(x_i)$ minimizing the loss, i.e.\ to minimize the following 
loss:%
\begin{equation}
  \mathcal{L}_{\textrm{DTI}}(c_{1:K},f_{1:K})= \sum_{i=1}^N 
  l(x_i,\,\{\cT_{f_1(x_i)}(c_1),
  \ldots,\cT_{f_K(x_i)}(c_K)\}),
    \label{eq:dti}
\end{equation}
where predictors $f_{1:K}$ are now shared for all samples. We found that using deep 
parameters predictors not only enables more efficient training but also leads to better 
clustering results especially with more complex transformations. Indeed, the structure and 
optimization of the predictors naturally regularize the parameters for each sample, without 
requiring any specific regularization loss, especially in the case of high numbers $N$ of 
samples and transformation parameters.

\vspace{-0.3em}
In the next section we present concrete losses and algorithms. We then describe 
differentiable modules for relevant transformations and discuss parameter predictor 
architecture as well as training in Section~\ref{sec:learning}. 

\subsection{Application to K-means and GMM}

\paragraph{K-means.}
The goal of K-means algorithm~\cite{macqueenMethodsClassificationAnalysis1967} is to find a 
set of prototypes $c_{1:K}$ such that the average Euclidean distance between each sample and 
the closest prototype is minimized. Following the reasoning of Section~\ref{sec:dti}, the 
loss optimized in K-means can be transformed into a transformation-invariant loss:%
\begin{equation}
  \mathcal{L}_{\textrm{DTI K-means}}(c_{1:K},f_{1:K})= \sum_{i=1}^N   
  \underset{k}{\min}~\|x_i - \cT_{f_k(x_i)}(c_k)\|^2.
  \label{eq:dtikmeans}
\end{equation}
Following batch gradient-based trainings~\cite{bottouConvergencePropertiesKmeans1995} of 
K-means, we can then simply jointly minimize $\cL_{\textrm{DTI K-means}}$ over prototypes 
$c_{1:K}$ and deep transformation parameter predictors $f_{1:K}$ using a batch gradient 
descent algorithm. In practice, we initialize prototypes $c_{1:K}$ with random samples and 
predictors $f_{1:K}$ such that $\forall k,\forall x,\cT_{f_k(x)} = \id$.
\begin{algorithm}[t]
  \caption{Deep Transformation-Invariant Gaussian Mixture Model}
  \setlength{\abovedisplayskip}{0pt}
  \setlength{\belowdisplayskip}{0pt}
  \KwIn{data $\mathrm{\mathbf{X}}$, number of clusters $K$, transformation $\cT$}
  \KwOut{cluster assignations, Gaussian parameters $\mu_{1:K}, \Sigma_{1:K}$, deep predictors 
  $f_{1:K}$}
  \KwInit{$\mu_{1:K}$ with random samples, $\Sigma_{1:K} = 0.5$, $\eta_{1:K} = 1$
  and $\forall k,\forall x,\cT_{f_k(x)} = \id$}
  \While{not converged}{\
    i.\ sample a batch of data points $x_{1:N}$\\
    ii.\ compute mixing probabilities:\quad$\pi_{1:K} = \softmax(\eta_{1:K})$\\
    iii.\ compute per-sample Gaussian transformed parameters:$$\forall k,~\forall 
    i,\;\;\tilde{\mu}_{ki} = \cT_{f_k(x_i)}(\mu_k)\;\;\mathrm{and}\;\;\tilde{\Sigma}_{ki} = 
    \cT_{f_k(x_i)}^{*}(\Sigma_k) + \diag(\sigma_{\mathrm{min}}^2)$$\\
    iv.\ compute responsibilities:
      $\forall k,~\forall i,~\gamma_{ki} = \frac{\pi_k 
      G(x_i\,;\tilde{\mu}_{ki},\tilde{\Sigma}_{ki})}
      {\sum_{j} \pi_{j} G(x_i\,;\tilde{\mu}_{ji},\tilde{\Sigma}_{ji})}$
      \hfill{(\textit{E-step})}\\
    v.\ minimize expected negative log-likelihood w.r.t to $\{\mu_{1:K}, \Sigma_{1:K}, 
    \eta_{1:K}, f_{1:K}\}$:
    \begin{equation}\tag{\textit{M-step}}
      \mathbb{E}[\cL_{\textrm{DTI GMM}}] = -\sum_{i=1}^{N} \sum_{k=1}^K \gamma_{ki}\Big(
      \log\big(G(x_i\,;\tilde{\mu}_{ki},\tilde{\Sigma}_{ki})\big) + \log(\pi_k)\Big)
    \end{equation}
  }
  \label{algo:dtigmm}
\end{algorithm}
\vspace{-0.7em}
\paragraph{Gaussian mixture model.}
We now consider that data are observations of a mixture of $K$ multivariate normal random 
variables $X_{1:K}$, i.e. $X = \sum_{k} \delta_{k, \Delta}X_k$ where $\delta$ is the 
Kronecker function and $\Delta \in \{1,\ldots,K\}$ is a random variable defined by $P(\Delta 
= k) = \pi_k$, with $\forall k,\,\pi_k>0$ and $\sum_k\pi_k = 1$. We write $\mu_k$ and 
$\Sigma_k$ the mean and covariance of $X_k$
and $G(\,\textbf{.}\,;\mu_k,\Sigma_k)$ associated probability density function.  The 
transformation-invariant negative log-likelihood can then be written:
\begin{equation}
  \mathcal{L}_{\textrm{DTI GMM}}(\mu_{1:K}, \Sigma_{1:K}, \pi_{1:K}, f_{1:K}) = 
  -\sum_{i=1}^{N} \log\Big(\sum_{k=1}^K\pi_k G\big(x_i \,; 
  \scaleto{\cT_{f_k(x_i)}}{10pt}(\mu_{k}), 
  \scaleto{\cT^{*}_{f_k(x_i)}}{11pt}(\Sigma_{k})\big)\Big),
  \label{eq:dtigmm}
\end{equation}
where $\cT^{*}$ is slightly modified version of $\cT$. Indeed, $\cT$ may include 
transformations that one can apply to the covariance, such as spatial transformations, and 
other that would not make sense, such as additive color transformations. We jointly minimize 
$\mathcal{L}_{\textrm{DTI GMM}}$ over Gaussian parameters, mixing probabilities, and deep 
transformation parameters $f_{1:K}$ using a batch gradient-based EM procedure similar 
to~\cite{hosseiniMatrixManifoldOptimization2015,greffNeuralExpectationMaximization2017, 
gepperthGradientbasedTrainingGaussian2019} and detailed in Algorithm~\ref{algo:dtigmm}.
In practice, we assume that pixels are independent resulting in diagonal covariance matrices.  

In such gradient-based procedures, two constraints have to be enforced, namely the 
positivity and normalization of mixing probabilities $\pi_k$ and the non-negativeness of the
diagonal covariance terms.  For the mixing probabilities constraints, we adopt the approach 
used in~\cite{hosseiniMatrixManifoldOptimization2015} 
and~\cite{gepperthGradientbasedTrainingGaussian2019} which optimize mixing parameters 
$\eta_k$ used to compute the probabilities $\pi_k$ using a softmax instead of directly 
optimizing $\pi_k$, which we write $\pi_{1:K} = \softmax(\eta_{1:K})$. For the variance 
non-negativeness, we introduce a fixed minimal variance value $\sigma_{\mathrm{min}}^2$ which 
is added to the variances when evaluating the probability density function. This approach is 
different from the one in~\cite{gepperthGradientbasedTrainingGaussian2019} which instead use 
clipping, because we found training with clipped values was harder. In practice, we take 
$\sigma_{\textrm{min}} = 0.25$.

\section{Learning image transformations}\label{sec:learning}

\subsection{Architecture and transformation modules}
\label{sec:transfo}
We consider a set of prototypes $c_{1:K}$ we would like to transform to match a given sample 
$x$. To do so, we propose to learn for each prototype $c_k$, a separate deep predictor which 
predicts transformation parameters $\beta$. We propose to model the family of transformations  
$\cT_{\beta}$ as a sequence of M parametric transformations such that, writing 
$\beta=(\beta^1, \ldots,\beta^M)$, $\cT_{\beta} = \cT^{M}_{\beta^M} \circ \dotsc \circ 
\cT^{1}_{\beta^1}$. In the following, we describe the architecture of transformation 
parameter predictors $f_{1:K}$, as well as each family of parametric transformation modules 
we use.  Figure~\ref{fig:method_b} shows our learned transformation process on a MNIST 
example.

\vspace{-0.7em}
\paragraph{Parameters prediction network.} For all experiments, we use the same parameter 
predictor network architecture composed of a shared ResNet~\cite{heDeepResidualLearning2016}
backbone truncated after the global average pooling, followed by $K\times M$ Multi-Layer 
Perceptrons (MLPs), one for each prototype and each transformation module. For the ResNet 
backbone, we use ResNet-20 for images smaller than $64\times 64$ and ResNet-18 otherwise. 
Each MLP has the same architecture, with two hidden layers of 128 units.

\vspace{-0.7em}
\paragraph{Spatial transformer module.}  To model spatial transformations of the prototypes, 
we follow the spatial transformers developed by Jaderberg et  
al.~\cite{jaderbergSpatialTransformerNetworks2015}. The key idea is to model spatial 
transformations as a differentiable image sampling of the input using a deformed sampling 
grid. We use affine $\cT^{\,\textrm{aff}}_\beta$, projective $\cT^{\,\textrm{proj}}_\beta$ 
and thin plate spline 
$\cT^{\,\textrm{tps}}_\beta$~\cite{booksteinPrincipalWarpsThinplate1989} transformations 
which respectively correspond to 6, 8 and 16 (a 4x4 grid of control points) parameters.

\vspace{-0.7em}
\paragraph{Color transformation module.} We model color transformation with a channel-wise 
diagonal affine transformation on the full image, which we write 
$\cT^{\,\textrm{col}}_\beta$. It has 2 parameters for greyscale images and 6 parameters for 
colored images. We first used a full affine transformation with 12 parameters, however the 
network was able to hide several patterns in the different color channels of a single 
prototype (Appendix~\ref{sec:color_discuss}). Note that a similar transformation was 
theoretically introduced in capsules~\cite{kosiorekStackedCapsuleAutoencoders2019}, but with 
the different goal of obtaining a color-invariant feature representation. Deep feature-based 
approaches often handle color images with a pre-processing step such as Sobel 
filtering~\cite{caronDeepClusteringUnsupervised2018, jiInvariantInformationClustering2019, 
kosiorekStackedCapsuleAutoencoders2019}.
We believe the way we align colors of the prototypes to obtain color invariance in pixel 
space is novel, and it enables us to directly work with colored images without using any 
pre-processing or specific invariant features.

\vspace{-0.7em}
\paragraph{Morphological transformation module.} We introduce a new transformation module to 
learn morphological operations~\cite{haralick1987image} such as dilation and erosion. We 
consider a greyscale image $x \in \RR^{D}$ of size $U\times V=D$, we write $x[u,v]$ the value 
of the pixel $(u,v)$ for $u\in \{1,\ldots,U\}$ and $v\in \{1,\ldots,V\}$. Given a 2D region 
$A$, the dilation of $x$ by $A$, $\mathcal{D}_A(x)\in \RR^{D}$, is defined by 
$\mathcal{D}_A(x)[u,v]=\max_{(u',v') \in A}~x[u+u',v+v']$ and its erosion by $A$, 
$\mathcal{E}_A(x)\in \RR^{D}$, is defined by $\mathcal{E}_A(x)[u,v]=\min_{(u',v') \in 
A}~x[u+u',v+v']$. Directly learning the region $A$ which parametrizes these transformations 
is challenging, we thus propose to learn parameters $(\alpha, a)$ for the following soft 
version of these transformations:
\begin{equation}
  \cT^{\,\textrm{mor}}_{(\alpha, a)}(x)[u,v] = \dfrac{\sum_{(u',v') \in W}~x[u+u',v+v']\cdot 
  a[u+u',v+v'] \cdot e^{\alpha x[u+u',v+v']}  }{\sum_{(u',v') \in W}~a[u+u',v+v']\cdot 
  e^{\alpha x[u+u',v+v']}},
\end{equation}

where $W$ is a fixed set of 2D positions, $\alpha$ is a softmax (positive values) or softmin 
(negative values) parameter and $a$ is a set of parameters with values between 0 and 1 
defined for every position $(u',v')\in W$. Parameters $a$ can be interpreted as an image, or 
as a soft version of the region $A$ used for morphological operations. Note that if 
$a[u',v']=\mathbf{1}_{\{(u',v')\in A\}}$, when $\alpha \rightarrow +\infty$ (resp.  
$-\infty$), it successfully emulates $\mathcal{D}_A$ (resp. $\mathcal{E}_A$). In practice, we 
use a grid of integer positions around the origin of size $7\times 7$ for $W$. Note that 
since morphological transformations do not form a group, transformation-invariant 
denomination is slightly abusive.

\subsection{Training}\label{sec:training}

We found that two key elements were critical to obtain good results: empty cluster 
reassignment and curriculum learning. We then discuss further implementation details and 
computational cost.

\vspace{-0.7em}
\paragraph{Empty cluster reassignment.} Similar 
to~\cite{caronDeepClusteringUnsupervised2018}, we adopt an empty cluster reassignment 
strategy during our clustering optimization. We reinitialize both prototype and deep 
predictor of "tiny" clusters using the parameters of the largest cluster with a small added 
noise. In practice, the size of balanced clusters being $N/K$, we define "tiny" as less than 
20\% of $N/K$.

\vspace{-0.7em}
\paragraph{Curriculum learning.} Learning to predict transformations is a hard task, 
especially when the number of parameters is high. To ease learning, we thus adopt a 
curriculum learning strategy by gradually adding more complex transformation modules to the 
training.  Given a target sequence of transformations to learn, we first train our model 
without any transformation - or equivalently with an identity module - then iteratively add 
subsequent modules once convergence has been reached. We found this is especially important 
when modeling local deformations with complex transformations with many parameters, such as 
TPS and morphological transformations.  Intuitively, prototypes should first be coarsely 
aligned before attempting to refine the alignment with more complex transformations.

\vspace{-0.7em}
\paragraph{Implementation details.} Both clustering parameters and parameter prediction 
networks are learned jointly and end-to-end using Adam 
optimizer~\cite{kingmaAdamMethodStochastic2015} with a 10$^{-6}$ weight decay on the neural 
network parameters. We sequentially add transformation modules at a constant learning rate of 
0.001 then divide the learning rate by 10 after convergence - corresponding to different 
numbers of epochs depending on the dataset characteristics - and train for a few more epochs 
with the smaller learning rate. We use a batch size of 64 for real photograph collections and 
128 otherwise.

\vspace{-0.7em}
\paragraph{Computational cost.} Training DTI K-means or DTI GMM on MNIST takes approximately 
50 minutes on a single Nvidia GeForce RTX 2080 Ti GPU and full dataset inference takes 30 
seconds. We found it to be much faster than directly optimizing transformation parameters (TI 
clustering) for which convergence took more than 10 hours of training.

\section{Experiments}\label{sec:results}

\begin{table}
  \renewcommand{\arraystretch}{1}
  \addtolength{\tabcolsep}{-4pt}
  \caption{\textbf{Comparisons.}
    We report ACC and NMI in \% on standard clustering benchmarks.  Symbols mark methods that 
    use data augmentation ($\triangledown$) and manually selected features as input 
    ($\mathsection$ for pretrained features from best VaDE run, $\dagger$ for GIST features, 
  $\ddagger$ for Sobel filters) and are thus not directly comparable. For SVHN, we also 
  report our results with our Gaussian weighted loss ($\star$). Eval column refers to the 
  aggregate used: best run (\textit{max}), average (\textit{avg}) or run with minimal loss 
  (\textit{minLoss}).
}
  \vspace{0.2em}
  \centering
  \scriptsize
  \begin{tabular}{@{}lccccccccccccc@{}} \toprule
  &  & & \multicolumn{2}{c}{MNIST} & \multicolumn{2}{c}{MNIST-test} &
  \multicolumn{2}{c}{USPS} & \multicolumn{2}{c}{F-MNIST} & \multicolumn{2}{c}{FRGC} & SVHN\\
  \cmidrule(lr){4-5}  \cmidrule(lr){6-7}  \cmidrule(lr){8-9} \cmidrule(lr){10-11} 
  \cmidrule(lr){12-13} \cmidrule(ll){14-14}
  Method & Runs & Eval & ACC & NMI & ACC & NMI & ACC & NMI & ACC & NMI & ACC & NMI & ACC\\
  \midrule
  \multicolumn{13}{l}{\textit{Clustering on a learned feature}} \\
  \quad DEC~\cite{xieUnsupervisedDeepEmbedding2016, yangDeepSpectralClustering2019} & 9 & max 
  & 86.3 & 83.4 & 85.6 & 83.0 & 76.2 & 76.7 & 51.8 & 54.6 & 37.8 & 50.5 & - \\
  \quad InfoGAN~\cite{chenInfoGANInterpretableRepresentation2016, 
  mukherjeeClusterGANLatentSpace2019} & 5 & max &
  89.0 & 86.0 & - & - & - & - & 61.0 & 59.0 & - & - & - \\
  \quad VaDE~\cite{jiangVariationalDeepEmbedding2017, yangDeepSpectralClustering2019} & 10 & 
  max & 94.5 & 87.6 & - & - & 56.6 & 51.2 & 57.8 & 63.0 & - & - & - \\
  \quad ClusterGAN~\cite{mukherjeeClusterGANLatentSpace2019} & 5 & max &
  95.0 & 89.0 & - & - & - & - & 63.0 & 64.0 & - & - & - \\
  \quad JULE~\cite{yangJointUnsupervisedLearning2016} & 3 & avg &
  96.4 & 91.3 & 96.1 & 91.5 & 95.0 & 91.3 & 56.3 & 60.8 & 46.1 & 57.4 & -\\
  \quad DEPICT~\cite{dizajiDeepClusteringJoint2017} & 5 & avg &
  96.5 & 91.7 & 96.3 & 91.5 &\bf 96.4 &\bf 92.7 & 39.2 & 39.2 &\bf 47.0 &\bf 61.0 & - \\
  \quad DSCDAN~\cite{yangDeepSpectralClustering2019} & 10 & avg & \bf 97.8 &\bf 94.1 &\bf 
  98.0 & \bf 94.6 & 86.9 & 85.7 &\bf 66.2 &\bf 64.5 & - & - & - \\ \hdashline
  \multicolumn{13}{l}{\textit{Clustering on a learned feature with data augmentation and/or 
  ad hoc data representation}}\\
  \quad SpectralNet~\cite{shahamSpectralNetSpectralClustering2018} & 5 & avg &
  97.1$^\mathsection$ &\bf 92.4$^\mathsection$ & - & - & - & - & - & - & - & - & - \\
  \quad IMSAT~\cite{huLearningDiscreteRepresentations2017} & 12 & avg &
  98.4$^{\triangledown}$ & - &
  - & - & - & - & - & - & - & - & \bf 57.3$^{\triangledown\dagger}$\\
  \quad ADC~\cite{hausserAssociativeDeepClustering2018} & 20 & avg &
  98.7$^{\triangledown}$ & - & - & - & - & - & - & - &\bf 43.7$^{\triangledown}$ & - & 
  38.6$^{\triangledown}$\\
  \quad SCAE~\cite{kosiorekStackedCapsuleAutoencoders2019} & 5 & avg &
  98.7$^{\triangledown}$ & - & - & - & - & - & - & - & - & - & 55.3$^{\ddagger}$\\
  \quad IIC~\cite{jiInvariantInformationClustering2019} & 5 & avg &
  98.4$^{\triangledown}$ & - & -  & - & - & - & - & - & - & - & -\\
  & 5 & minLoss &\bf 99.2$^{\triangledown}$ & - & - & - & - & - & - & - & - & - & -\\

  \midrule
  \multicolumn{13}{l}{\textit{Clustering on pixel values}}\\
  \quad K-means~\cite{macqueenMethodsClassificationAnalysis1967} & 10 & avg &
  54.8 & 50.2 & 55.9 & 51.2 & 65.3 & 61.2 & 54.1 & 51.4 & 22.7 & 26.5 & 12.2\\
  \quad GMM~\cite{dempsterMaximumLikelihoodIncomplete1977} & 10 & avg &
  54.2 & 51.7 & 55.6 & 54.7 & 66.0 & 60.9 & 49.7 & 51.2 & 24.2 & 27.9 & 11.6\\
  \quad \textbf{DTI K-means}
  & 10 & avg & \bf 97.3 & \bf 94.0 & 96.6  & 94.6 & 86.4 & 88.2 & 61.2 & 63.7 & 39.6 & 48.7 & 
  36.4 / 44.5$^{\star}$\\
  & 10 & minLoss & 97.2 & 93.8 &\bf 98.0 &\bf 95.3 &\bf 89.8 &\bf 89.5 & 57.4 & 64.1 & 41.1 & 
  49.7  & 39.6 / 62.6$^{\star}$\\
  \quad \textbf{DTI GMM}
  & 10 & avg & 95.9 & 93.2 & 97.8 & 94.7 & 84.5 & 87.2 & 59.6 & 62.2 & 40.1 & 48.9 & 36.7 / 
  57.4$^{\star}$\\
  & 10 & minLoss & 97.1 & 93.7 &\bf 98.0 & 95.1 & 87.3 & 89.0 &\bf 68.2 &\bf 66.3 &\bf 41.6  
  &\bf 51.1 & 39.5 / \textbf{63.3}$^{\star}$\\
  \bottomrule
  \end{tabular}
  \label{tab:bench}
\end{table}

In this section, we first analyze our approach and compare it to state-of-the-art, then 
showcase its interest for image collection analysis and visualization. 

\subsection{Analysis and comparisons}

Similar to previous work on image clustering, we evaluate our approach with global 
classification accuracy (ACC), where a cluster-to-class mapping is computed using the 
Hungarian algorithm~\cite{kuhnHungarianMethodAssignment1955}, and Normalized Mutual 
Information (NMI). Datasets and corresponding transformation modules we used are described in 
Appendix~\ref{sec:dataset}.%
\begin{figure}
    \centering
    \begin{subfigure}{\columnwidth}
      \centering
      \includegraphics[width=\columnwidth]{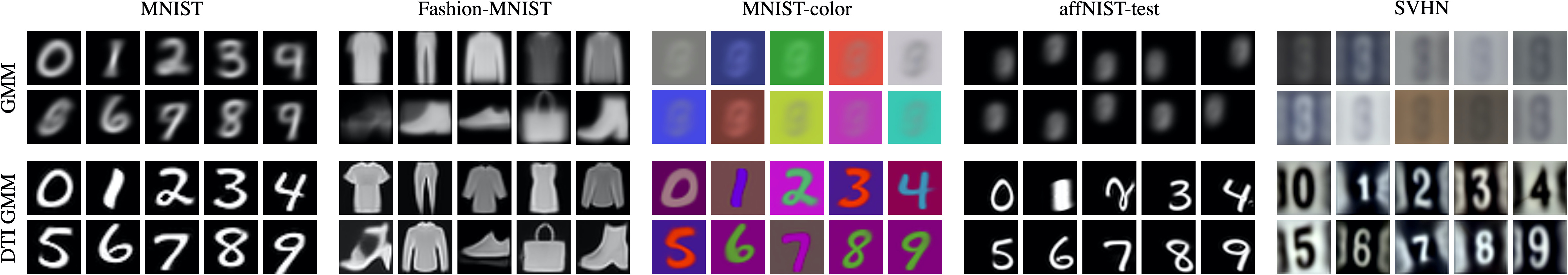}
      \caption{Prototypes learned for different datasets}
    \label{fig:cluster_centers}
    \end{subfigure}
    \begin{subfigure}{\columnwidth}
      \centering
      \vspace{0.1em}
      \includegraphics[width=\columnwidth]{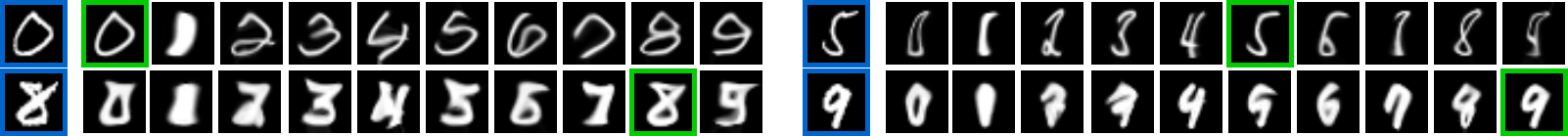}
      \caption{Transformations predicted for all prototypes for 4 MNIST images}
    \label{fig:transformations}
    \end{subfigure}
    \caption{\textbf{Qualitative results.} \textbf{(a)} compares prototypes learned from GMM 
    and our DTI GMM, \textbf{(b)} shows transformed prototypes given \tsfblue{query samples} 
  from MNIST and highlight the \tsfgreen{closest prototype}.}
  \label{fig:qualitative_results}
  \vspace{-1.5em}
\end{figure}

\vspace{-0.7em}
\paragraph{Comparison on standard benchmarks.}
In Table~\ref{tab:bench}, we report our results on standard image clustering benchmarks, 
i.e.\  digit datasets (MNIST~\cite{lecunGradientBasedLearningApplied1998}, 
USPS~\cite{hastie01statisticallearning}), a clothing dataset 
(Fashion-MNIST~\cite{xiao2017fashion}) and a face dataset (FRGC~\cite{FRGC}). We also
report results for SVHN~\cite{netzer2011reading} where concurrent methods use pre-processing 
to remove color bias. In the table, we separate representation-based from pixel-based methods 
and mark results using data augmentation or manually selected features as input. Note that 
our results depend on initialization, we provide detailed statistics in 
Appendix~\ref{sec:statistics}.

\vspace{-0.3em}
Our DTI clustering is fully unsupervised and does not require any data augmentation, ad hoc 
features, nor any hyper-parameter while performing clustering directly in pixel space.  We 
report average performances and performances of the minimal loss run which we found to 
correlate well with high performances (Appendix~\ref{sec:correl}). Because this non-trivial 
criterion allows to automatically select a run in a fully unsupervised way, we argue it can 
be compared to average results from competing methods which don't provide such criterion.
 
\vspace{-0.3em}
First, DTI clustering achieves competitive results on all datasets, in particular improving
state-of-the-art by a significant margin on SVHN and Fashion-MNIST.\ For SVHN, we first found 
that the prototypes quality was harmed by digits on the side of the image. To pay more 
attention to the center digit, we weighted the clustering loss by a Gaussian weight 
($\sigma=7$). It led to better prototypes and allowed us to improve over all concurrent 
methods by a large margin. Compared to representation-based methods, our pixel-based 
clustering is highly interpretable.  Figure~\ref{fig:cluster_centers} shows standard GMM 
prototypes and our prototypes learned with DTI GMM which appear to be much sharper than 
standard ones. This directly stems from the quality of the learned transformations, 
visualized in Figure~\ref{fig:transformations}.  Our transformation modules can successfully 
align the prototype, adapt the thickness and apply local elastic deformations. More alignment 
results are available on our project 
\href{http://imagine.enpc.fr/~monniert/DTIClustering/}{webpage}.


\begin{table}
  \centering
  \begin{minipage}{0.55\linewidth}
  \small
  \caption{\textbf{Augmented and specific datasets.} Clustering accuracy (\%) with standard 
  deviation for methods applied on raw images (no pre-processing). We used 10 runs for our 
method and 5 for the baselines.}
  \centering
  \scriptsize
  \begin{tabular}{@{}lcccc@{}} \toprule
  Method & Eval &  MNIST-1k & MNIST-color & affNIST-test\\
  \midrule
  VaDE~\cite{jiangVariationalDeepEmbedding2017}  & avg  & 49.6 (5.6) & 11.9 (1.2) & Div.\\
  IMSAT~\cite{huLearningDiscreteRepresentations2017}
  & avg & 67.9 (2.3) & 10.6 (0.1) & 18.2 (2.6)  \\ 
  IIC~\cite{jiInvariantInformationClustering2019}
  & avg & 63.4 (0.4) & 10.6 (0.0) &57.6 (0.0)\\
  & minLoss  & 63.2 & 10.6 & 57.6\\
  \midrule
  \textbf{DTI K-means} & avg & 79.8 (6.9) & 96.7 (0.1) & 95.5 (3.3) \\
                       &  minLoss & \bf 90.5 &\bf 96.8 &\bf 97.0\\
  \textbf{DTI GMM} &  avg & 80.8 (7.2)  & 96.0 (0.2) & 93.3 (5.9)\\
                   &  minLoss & 87.1 & 95.8 &\bf 97.0\\
  \bottomrule
  \end{tabular}
  \label{tab:syn_bench}
  \end{minipage}\quad\enspace
  \begin{minipage}{0.4\linewidth}
  \small
  \caption{\textbf{Ablation study on MNIST.} Clustering accuracy (\%) over 10 runs.}
  \centering
  \scriptsize
  \vspace{0.54em}
  \begin{tabular}{@{}lcc@{}} \toprule
  Method &  Avg & MinLoss \\
  \midrule
  \bf DTI clustering (aff-morpho-tps) &\bf 97.3 & \bf 97.2 \\
  \quad ordering: aff-tps-morpho  & 95.5& 96.9 \\
  \quad ordering: morpho-aff-tps & 27.5& 97.0 \\
  \quad w/o morphological & 94.8  & 95.8 \\
  \quad w/o thin plate spline & 90.0 & 82.5\\
  \quad w/o affine & 85.1   & 96.8 \\
  \quad affine only & 90.1   & 90.5\\
  \quad w/o empty cluster reassignment& 80.9 & 78.6 \\
  \quad w/o curriculum learning & 83.9 & 78.9 \\
  TI clustering (aff-morpho-tps, 1 run) & 26.3 & 26.3 \\
  TI clustering (affine only) &73.0 & 73.1 \\
  \bottomrule
  \end{tabular}
  \label{tab:ablation}
  \end{minipage}
  \vspace{-0.5em}
\end{table}

\vspace{-0.7em}
\paragraph{Augmented and specific datasets.}
DTI clustering also works on small, colored and misaligned datasets. In 
Table~\ref{tab:syn_bench}, we highlight these strengths on specifics datasets generated from 
MNIST: MNIST-1k is a 1000 images subset, MNIST-color is obtained by randomly selecting a 
color for the foreground and background and 
affNIST-test\footnote{https://www.cs.toronto.edu/\textasciitilde tijmen/affNIST/} is the 
result of random affine transformations.  We used an online 
implementation\footnote{https://github.com/GuHongyang/VaDE-pytorch} for 
VaDE~\cite{jiangVariationalDeepEmbedding2017} and official ones for 
IMSAT~\cite{huLearningDiscreteRepresentations2017} and 
IIC~\cite{jiInvariantInformationClustering2019} to obtain baselines.  Our results show that 
the performances of DTI clustering is barely affected by spatial and color transformations, 
while baseline performances drop on affNIST-test and are almost chance on MNIST-color.  
Figure~\ref{fig:cluster_centers} shows the quality and interpretability of our cluster 
centers on affNIST-test and MNIST-color. DTI clustering also seems more data-efficient than 
the baselines we tested.

\vspace{-0.7em}
\paragraph{Ablation on MNIST.}
In Table~\ref{tab:ablation}, we conduct an ablation study on MNIST of our full model trained 
following Section~\ref{sec:training} with affine, morphological and TPS transformations.
We first explore the effect of transformation modules. Their order is not crucial, as shown 
by similar minLoss performances, but can greatly affect the stability of the training, as can 
be seen in the average results. Each module contributes to the final performance, affine 
transformations being the most important. We then validate our training strategy showing that 
both empty cluster reassignment and curriculum learning for the different modules are 
necessary. Finally, we directly optimize the loss of Equation~\ref{eq:ti} (TI clustering) by 
optimizing the transformation parameters for each sample at each iteration of the batch 
clustering algorithm, without using our parameter predictors. With rich transformations which 
have many parameters, such as TPS and morphological ones, this approach fails completely.  
Using only affine transformations, we obtain results clearly superior to standard 
clustering, but worse than ours.

\begin{figure}
    \centering
    \begin{subfigure}{\columnwidth}
      \centering
      \includegraphics[width=0.49\columnwidth]{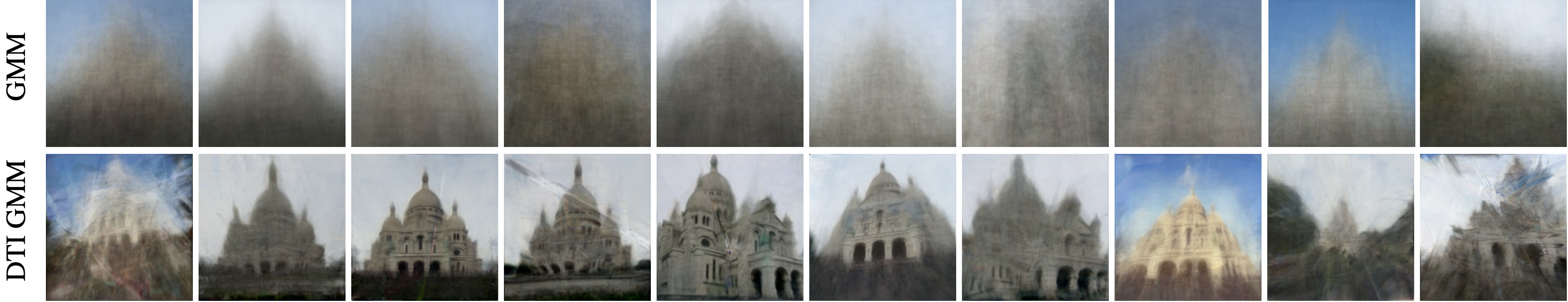}%
      \enspace\thinspace\thinspace
      \includegraphics[width=0.49\columnwidth]{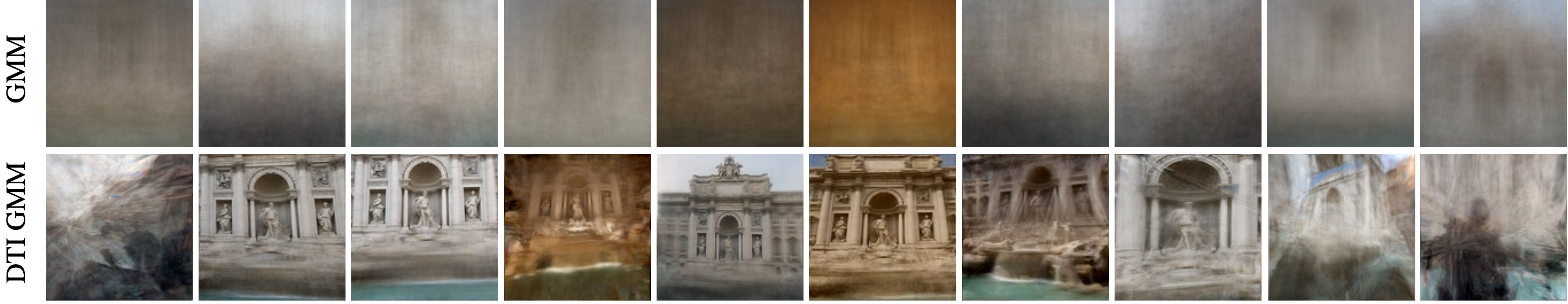}\\
      \vspace{0.2em}
      \includegraphics[width=0.49\columnwidth]{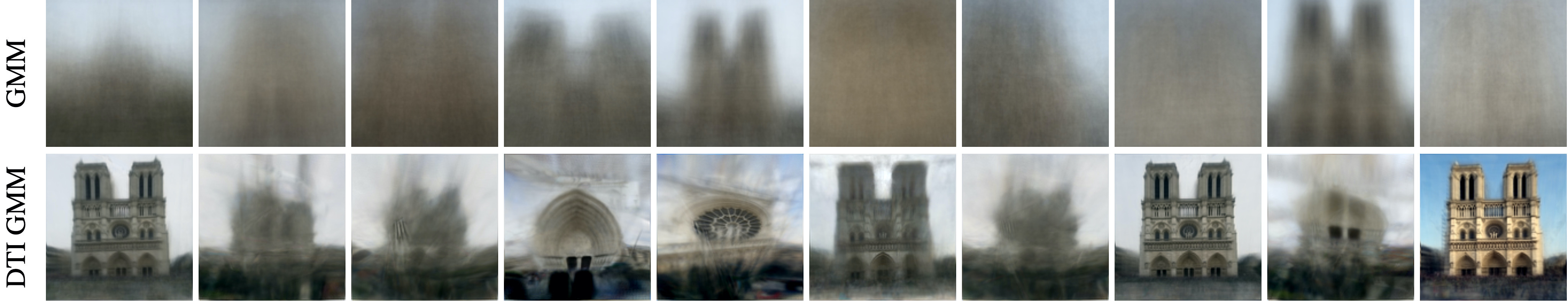}%
      \enspace\thinspace\thinspace
      \includegraphics[width=0.49\columnwidth]{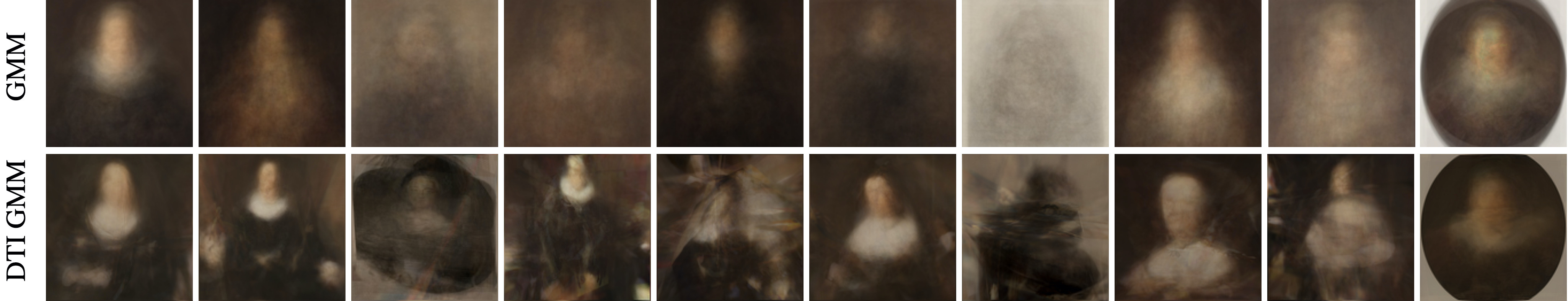}
      \caption{Full sets of prototypes discovered with GMM and DTI GMM}
    \end{subfigure}
    \begin{subfigure}{\columnwidth}
      \centering
      \vspace{0.1em}
      \includegraphics[width=\columnwidth]{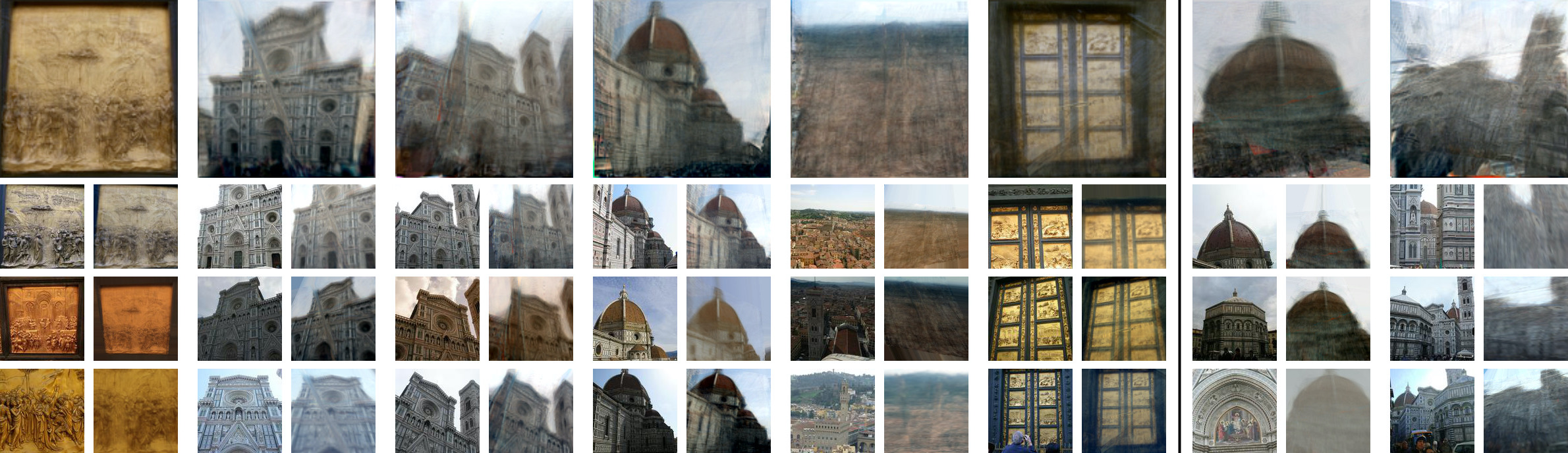}
      \caption{Examples of cluster centers and aligned images with DTI GMM (20 clusters)}
    \label{fig:flo}
    \end{subfigure}
      \caption{\textbf{Qualitative results on real photographs.} \textbf{(a)} Clustering 
        results from photographs of different locations in~\cite{li2018megadepth} (1,089 
        Sacre Coeur top-left, 1,688 Trevi fountain top-right, 2,625 Notre-Dame bottom-left) 
        and 980 Baroque portraits from~\cite{karayev2013recognizing} (bottom-right).  
        \textbf{(b)} Clustering results from 1,892 Florence cathedral images 
        from~\cite{li2018megadepth}.  Top row shows learned prototypes while the three bottom 
        rows show examples of images from each cluster and aligned prototypes. These clusters 
        contain respectively 44, 154, 134, 64, 71, 133, 85 and 64 images.  The left six 
      examples are successful clusters while the two right clusters are relative failure 
    cases.}
    \label{fig:real}
    \vspace{-1em}
\end{figure}

\subsection{Application to web images}
 
One of the main interest of our DTI clustering is that it allows to discover trends in real 
image collections. All images are resized and center cropped to 128$\times$128. The selection 
of the number of clusters is a difficult problem and is discussed in 
Appendix~\ref{sec:effect_k}.

\vspace{-0.3em}
In Figure~\ref{fig:teaser}, we show examples of prototypes discovered in very large 
unfiltered sets (15k each) of Instagram images associated to different 
hashtags\footnote{https://github.com/arc298/instagram-scraper was used to scrape photographs} 
using DTI GMM applied with 40 clusters. While many images are noise and are associated to 
prototypes which are not easily interpretable, we show prototypes where iconic photos and 
poses can be clearly identified. To the best of our knowledge, we believe we are the first to 
demonstrate this type of results from raw social network image collections. Comparable 
results in AverageExplorer~\cite{zhu2014averageExplorer}, e.g.\ on Santa images, could be 
obtained using ad hoc features and user interactions, while our results are produced fully 
automatically.

\vspace{-0.3em}
Figure~\ref{fig:real} shows qualitative clustering results on 
MegaDepth~\cite{li2018megadepth} and WikiPaintings~\cite{karayev2013recognizing}.
Similar to our results on image clustering benchmarks, our learned prototypes are more 
relevant and accurate than the ones obtained from standard clustering. Note that some of our 
prototypes are very sharp: they typically correspond to sets of photographs between which we 
can accurately model deformations, e.g.\ scenes that are mostly planar, with little 
perspective effects. On the contrary, more unique photographs and photographs with strong 3D 
effects that we cannot model will be associated to less interpretable and blurrier 
prototypes, such as the ones in the last two columns of Figure~\ref{fig:flo}. In 
Figure~\ref{fig:flo}, in addition to the prototypes discovered, we show examples of images 
contained in each cluster as well as the aligned prototype. Even for such complex images, the 
simple combination of our color and spatial modules manages to model real image 
transformations like illumination variations and viewpoint changes. More web image clustering 
results are shown on our project 
\href{http://imagine.enpc.fr/~monniert/DTIClustering/}{webpage}.
\vspace{-0.5em}

\section{Conclusion}
\vspace{-0.7em}
We have introduced an efficient deep transformation-invariant clustering approach in raw 
input space. Our key insight is the online optimization of a single clustering objective over 
clustering parameters and deep image transformation modules. We demonstrate competitive 
results on standard image clustering benchmarks, including improvements over state-of-the-art 
on SVHN and Fashion-MNIST.\ We also demonstrate promising results for real photograph
collection clustering and visualization. Finally, note that our DTI clustering framework is 
not specific to images and can be extended to other types of data as long as appropriate 
transformation modules are designed beforehand.
\vspace{-0.5em}

\section*{Acknowledgements}
\vspace{-0.7em}
This work was supported in part by ANR project EnHerit ANR-17-CE23-0008, project Rapid 
Tabasco, gifts  from  Adobe and HPC resources from GENCI-IDRIS (Grant 2020-AD011011697). We 
thank Bryan Russell, Vladimir Kim, Matthew Fisher, François Darmon, Simon Roburin, David 
Picard, Michaël Ramamonjisoa, Vincent Lepetit, Elliot Vincent, Jean Ponce, William Peebles 
and Alexei Efros for inspiring discussions and valuable feedback.

\newpage
\section*{Broader Impact}

The impact of clustering mainly depends on the data it is applied on. For instance, adding 
structure in user data can raise ethical concerns when users are assimilated to their 
cluster, and receive targeted advertisement and newsfeed. However, this is not specific to 
our method and can be said of any clustering algorithm. Also note that while our clustering  
can be applied for example to data from social media, the visual interpretation of the 
clusters it returns via the cluster centers respects privacy much better than showing 
specific examples from each cluster.

Because our method provides highly interpretable results, it might bring increased 
understanding of clustering algorithm results for the broader public, which we think may be a 
significant positive impact.

\small
\bibliography{references}

\newpage
\normalsize
\appendix

\section{Dataset descriptions}\label{sec:dataset}
\vspace{-1.5em}
\begin{table}[h!]
  \small
  \caption{\bf Datasets and transformation sequences used}
  \centering
  \addtolength{\tabcolsep}{-4pt}
  \begin{tabular}{@{}lcccc@{}}\toprule
  Dataset & Samples & Classes & Dimension & Transformation sequence\\\midrule
  \textit{Standard}\\
  \quad MNIST~\cite{lecunGradientBasedLearningApplied1998} & 70,000 & 10 & 
  1$\times$28$\times$28 & aff-morpho-tps\\
  \quad MNIST-test~\cite{lecunGradientBasedLearningApplied1998} & 10,000 & 10 & 
  1$\times$28$\times$28 & aff-morpho-tps\\
  \quad USPS~\cite{hastie01statisticallearning} & 9,298 & 10 & 1$\times$16$\times$16 & 
  col-aff-tps\\
  \quad Fashion-MNIST~\cite{xiao2017fashion} & 70,000 & 10 & 1$\times$28$\times$28 & 
  col-aff-tps\\
  \quad FRGC~\cite{FRGC} & 2,462 & 20 & 3$\times$32$\times$32 & col-aff-tps\\
  \quad SVHN~\cite{netzer2011reading}& 99,289 + unlabeled extra & 10 & 3$\times$28$\times$28 
  & col-proj\\
  \midrule
  \textit{Augmented}\\
  \quad MNIST-1k & 1,000 & 10 & 1$\times$28$\times$28 & aff-morpho-tps\\
  \quad MNIST-color & 70,000 & 10 & 3$\times$28$\times$28 & col-aff-tps\\
  \quad affNIST-test & 320,000 & 10 & 1$\times$40$\times$40 & aff-morpho-tps\\
  \midrule
  \textit{Real photographs}\\
  \quad All & 1k to 15k & - & 3$\times$128$\times$128 & col-proj\\
  \bottomrule
  \end{tabular}
  \label{tab:dataset}
\end{table}
Table~\ref{tab:dataset} summarizes dataset characteristics as well as the transformation 
sequences used. Datasets are:

\begin{itemize}[topsep=0pt,partopsep=0pt,parsep=0pt,leftmargin=7mm]
  \item \textbf{MNIST}  and {\bf MNIST-test}~\cite{lecunGradientBasedLearningApplied1998} 
    which respectively correspond to full and test subset of MNIST dataset. They depict 
    binary white handwritten digits centered over a black background.
  \item \textbf{USPS}~\cite{hastie01statisticallearning} is a handwritten digit dataset from  
    USPS composed of greyscale images.
  \item \textbf{Fashion-MNIST}~\cite{xiao2017fashion} is a 10-class 
    clothing dataset composed of greyscale images of cloth over black background. Classes 
    are: T-shirt, trouser, pullover, dress, coat, sandal, shirt, sneaker, bag, ankle boot.
  \item \textbf{FRGC}~\cite{FRGC} is a colored face dataset. We use a subset of this dataset
    introduced in~\cite{yangJointUnsupervisedLearning2016}, where 20 subjects are selected 
    and each image is cropped and resized to a constant size of 32$\times$32.
  \item \textbf{SVHN}~\cite{netzer2011reading} is composed of digits extracted from house 
    numbers cropped from Google Street View images. Following standard practice for
    clustering, we use both labeled samples (99,289) and unlabeled extra samples 
    (\textasciitilde 530k) for training and evaluate on the labeled subset only.
  \item \textbf{affNIST-test} is the test split of affNIST 
    (https://www.cs.toronto.edu/~tijmen/affNIST/) an augmented dataset of
    MNIST where random affine transformations are applied.
  \item \textbf{MNIST-1k}: we randomly sampled 1,000 images from the test split of MNIST.
  \item \textbf{MNIST-color}: we augmented MNIST with random colors for background and 
    foreground.
\end{itemize}

\section{Transformation invariance}\label{sec:proof}

We consider $N$ image samples $x_{1:N}$, $K$ prototypes $c_{1:K}$ and a group of parametric 
transformations $\{\cT_\beta,\, \beta\in B \}$.  For $\beta_1,\beta_2 \in B$, we write 
$\beta_1\beta_2 \in B$ the element such that $\cT_{\beta_1 \beta_2}=\cT_{\beta_1} \circ 
\cT_{\beta_2}$. We have, for any $\alpha_1, \ldots,\alpha_K \in B$:%
\begin{equation*}
  \mathcal{L}_{\textrm{TI}}(\{ c_1, \ldots , 
  c_K\})=\mathcal{L}_{\textrm{TI}}(\{\cT_{\alpha_1}(c_1), \ldots , \cT_{\alpha_K}(c_K)\}).
\end{equation*}
Indeed:\vspace{-0.5em}%
\begin{align*}
  \mathcal{L}_{\textrm{TI}}(\{\cT_{\alpha_1}(c_1),\ldots , 
  \cT_{\alpha_K}(c_K)\})&=\sum_{i=1}^N \min_{\{\beta_1,\ldots, \beta_K\} \in B^K}\  l(x_i,\,
  \{\cT_{\beta_1}\circ \cT_{\alpha_1}(c_1), \ldots,\cT_{\beta_K}\circ 
\cT_{\alpha_K}(c_K)\})\\
  &= \sum_{i=1}^N \min_{\{\beta_1,\ldots, \beta_K\} \in B^K}\  l(x_i,\,
  \{\cT_{\beta_1\alpha_1}(c_1), \ldots,\cT_{\beta_K\alpha_K}(c_K)\})\\
  &=\sum_{i=1}^N \min_{\{\beta'_1,\ldots, \beta'_K\} \in B^K}\  l(x_i,\,
  \{\cT_{\beta'_1}(c_1), \ldots,\cT_{\beta'_K}(c_K)\})\\
  &=\mathcal{L}_{\textrm{TI}}(\{ c_1,\ldots , c_K\}),
  \label{eq:tiproof}
\end{align*}
using the variable change $\beta'_k=\beta_k\alpha_k$, which is possible because for any 
$\alpha \in B$, $\alpha B = B$ as we assumed to have a group of transformations.

In some specific cases, the loss is also invariant to the samples, in particular when the 
loss $l$ is invariant to joint transformation of the prototype and the samples, i.e.\ for
any $\beta \in B$, $l(x_i,\{c_1,\ldots, c_K\}=l(\cT_\beta(x_i),\{\cT_\beta(c_1),\ldots,
\cT_\beta(c_K)\}$. This is the case for example for K-means with a group of isometric 
transformations (e.g.\ rigid transformations), and it is also the case for GMM with the group 
of affine transformations applied to both the mean and covariance mixture parameters.

Note that we also tried to transform the samples to match the prototypes, which would lead to 
an invariance to sample transformation. However, a trivial solution to corresponding 
optimization problem is to learn "empty" prototypes and transformations of the samples into 
empty images. For examples, for the MNIST case with affine transformations, we observed that 
completely black prototypes were learned and any digit was transformed into a black image.  
Although a regularization term could have prevented such behaviour, we argue that keeping raw 
samples as target and transforming the prototypes is simpler and effective.

\section{Analysis}\label{sec:analysis}

\subsection{Statistics on standard clustering benchmarks}\label{sec:statistics}
\vspace{-1.5em}
\begin{table}[!hb]
  \renewcommand{\arraystretch}{1}
  \addtolength{\tabcolsep}{-4pt}
  \caption{\textbf{Detailed results.} We report statistics of our results on standard 
  clustering benchmarks. For SVHN, we also report results with our Gaussian weighted loss 
($\star$).}
  \vspace{0.2em}
  \centering
  \scriptsize
  \begin{tabular}{@{}lccccccccccccc@{}} \toprule
  &  & & \multicolumn{2}{c}{MNIST} & \multicolumn{2}{c}{MNIST-test} &
  \multicolumn{2}{c}{USPS} & \multicolumn{2}{c}{F-MNIST} & \multicolumn{2}{c}{FRGC} & SVHN\\
  \cmidrule(lr){4-5}  \cmidrule(lr){6-7}  \cmidrule(lr){8-9} \cmidrule(lr){10-11} 
  \cmidrule(lr){12-13} \cmidrule(ll){14-14}
  Method & Runs & Stat & ACC & NMI & ACC & NMI & ACC & NMI & ACC & NMI & ACC & NMI & ACC\\
  \midrule

  \textbf{DTI K-means}
  & 10 & avg &  97.3 &  94.0 & 96.6  & 94.6 & 86.4 & 88.2 & 61.2 & 63.7 & 39.6 & 48.7 & 
  36.4 / 44.5$^{\star}$\\
  & 10 & std & 0.1 & 0.1 & 4.1 & 1.5 & 4.1 & 1.6 & 2.0 & 0.3 & 1.7 & 2.2 & 1.9 / 
  9.6$^{\star}$\\
  & 10 & min & 97.1 & 93.8 & 84.9 & 90.4 & 83.2 & 87.1 & 57.4 & 63.2 & 35.9 & 43.9 & 34.5 / 
  37.0$^{\star}$\\
  & 10 & median & 97.3 & 94.0 & 97.9 & 95.1 & 85.0 & 87.4 & 61.9 & 63.3 & 40.2 & 49.3 & 35.8 
  / 39.6$^{\star}$\\
  & 10 & max & 97.5 & 94.2 & 98.0 & 95.3 & 96.4 & 92.0 & 63.3 & 64.2 & 41.1 & 51.4 & 39.6 / 
  62.6$^{\star}$\\
  & 10 & minLoss & 97.2 & 93.8 & 98.0 & 95.3 & 89.8 & 89.5 & 57.4 & 64.1 & 41.1 & 49.7 & 39.6 
  / 62.6$^{\star}$\\

  \textbf{DTI GMM}
  & 10 & avg & 95.9 & 93.2 & 97.8 & 94.7 & 84.5 & 87.2 & 59.6 & 62.2 & 40.1 & 48.9 & 36.7 / 
  57.4$^{\star}$\\
  & 10 & std & 3.9 & 1.5 & 0.1 & 0.2 & 2.0 & 0.8 & 4.7 & 2.4 & 1.4 & 1.5 & 2.3 / 
  5.1$^{\star}$\\
  & 10 & min & 84.7 & 89.1 & 97.7 & 94.4 & 82.0 & 86.3 & 56.1 & 59.7 & 38.4 & 46.8 & 34.0 / 
  49.9$^{\star}$\\
  & 10 & median & 97.1 & 93.7 & 97.8 & 94.7 & 84.3 & 87.1 & 57.2 & 60.9 & 39.6 & 49.1 & 36.4 
  / 57.4$^{\star}$\\
  & 10 & max & 97.3 & 93.9 & 98.0 & 95.1 & 87.3 & 89.0 & 68.2 & 66.3 & 41.9 & 51.1 & 39.5 / 
  64.6$^{\star}$\\
  & 10 & minLoss & 97.1 & 93.7 & 98.0 & 95.1 & 87.3 & 89.0 & 68.2 & 66.3 & 41.6 & 51.1 & 39.5 
  / 63.3$^{\star}$\\
  \bottomrule
  \end{tabular}
  \label{tab:stats}
\end{table}

\subsection{Accuracy and loss correlation}\label{sec:correl}
\begin{wrapfigure}{r}{0.4\textwidth}
  \vspace{-1.3em}
  \centering
  \includegraphics[width=0.4\columnwidth]{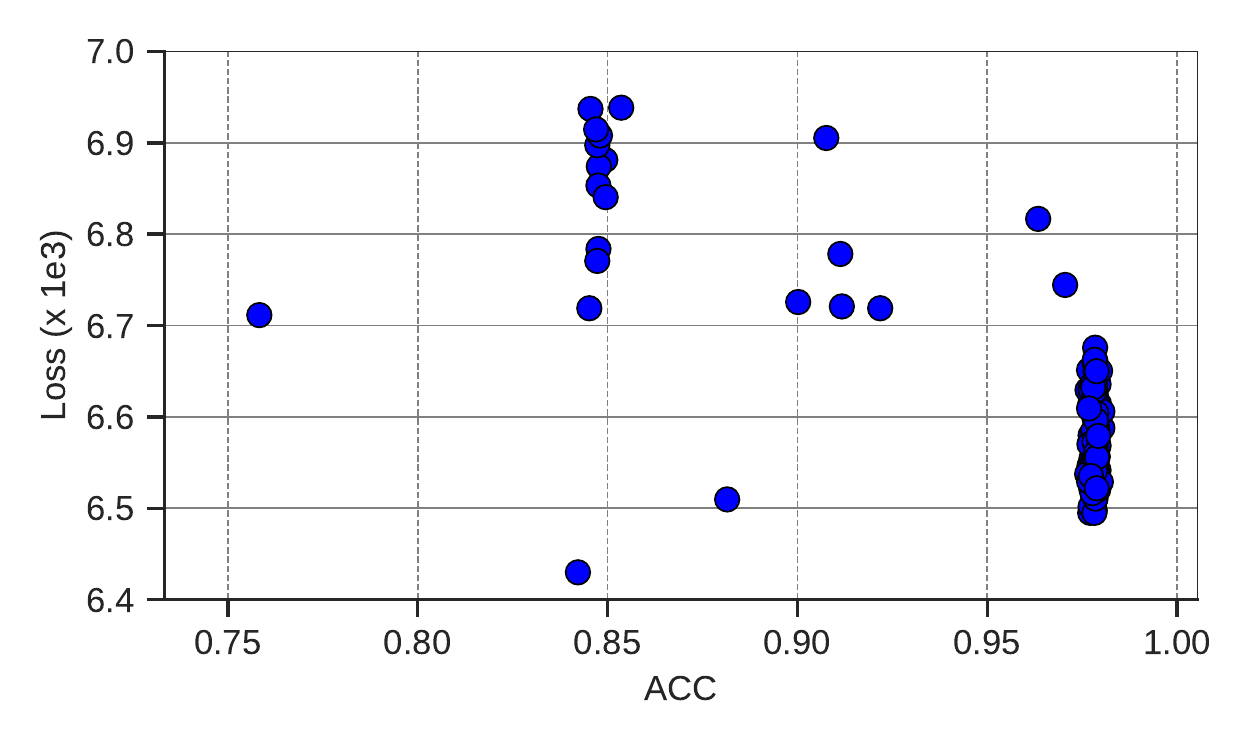}
  \vspace{-1em}
  \caption{\textbf{Accuracy/loss correlation.} We report loss and accuracy for DTI K-means on 
  MNIST-test.}
  \label{fig:correlation}
  \vspace{-1em}
\end{wrapfigure}
Similar to standard K-means and GMM, there is a variation in performances depending on the 
random initialization. We experimentally found that: (i) runs seem to be mainly grouped into 
distinct modes, each corresponding to roughly the same clustering quality; (ii) a run with a 
low loss usually leads to high clustering performances.  We launched 100 runs on MNIST-test 
dataset and plot the loss with respect to the accuracy for each run in 
Figure~\ref{fig:correlation}. Except 2 outliers for the 100 runs, the runs with lower loss 
correspond to the runs with better performances. This is verified in most of our experiments, 
where the minLoss criterion clearly improves over the average performance.

\subsection{Effect of the number of clusters K}\label{sec:effect_k}

\begin{figure}[!h]
  \centering
  \begin{subfigure}{\columnwidth}
    \centering
    \includegraphics[width=0.4\columnwidth]{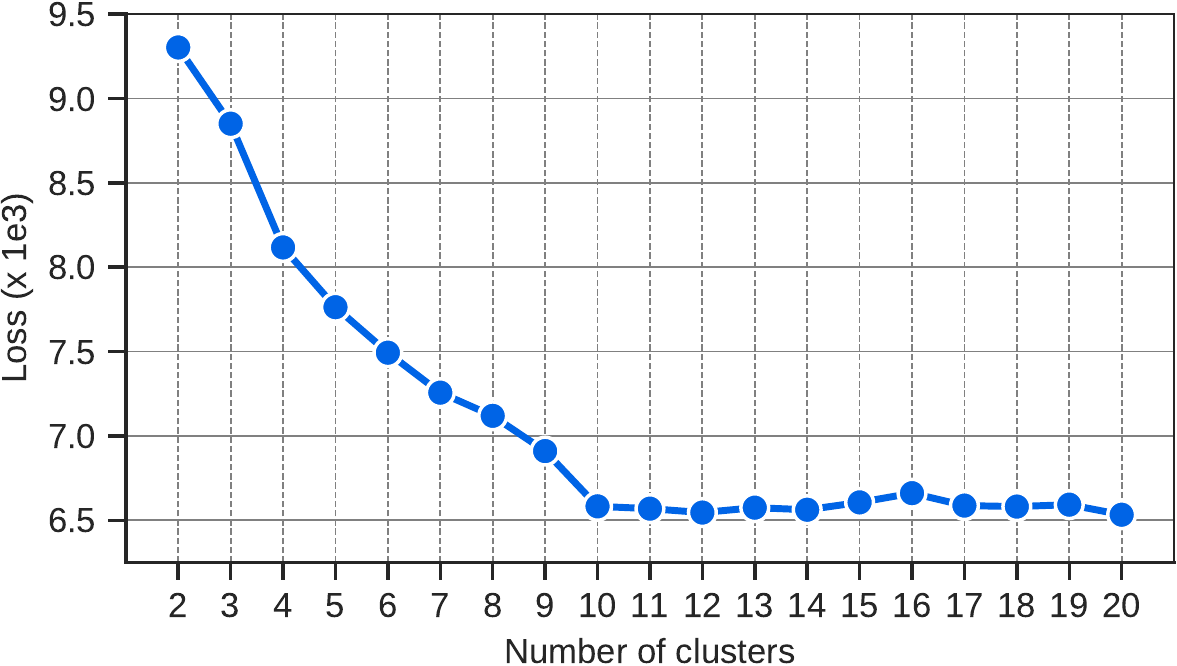}\quad
    \includegraphics[width=0.4\columnwidth]{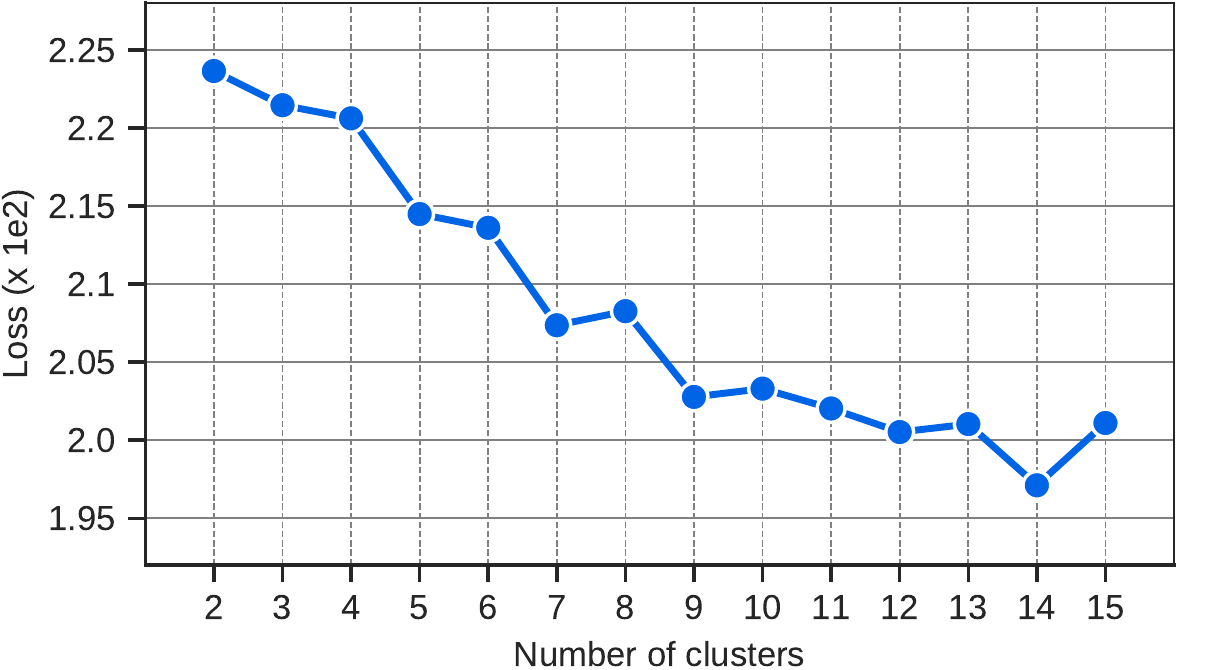}
    \caption{Loss w.r.t\ varying numbers of clusters for MNIST-test (left) and Notre-Dame 
    (right)}
    \label{fig:loss_k_quant}
  \end{subfigure}
  \begin{subfigure}{\columnwidth}
    \centering
    \vspace{0.1em}
    \includegraphics[width=0.9\columnwidth]{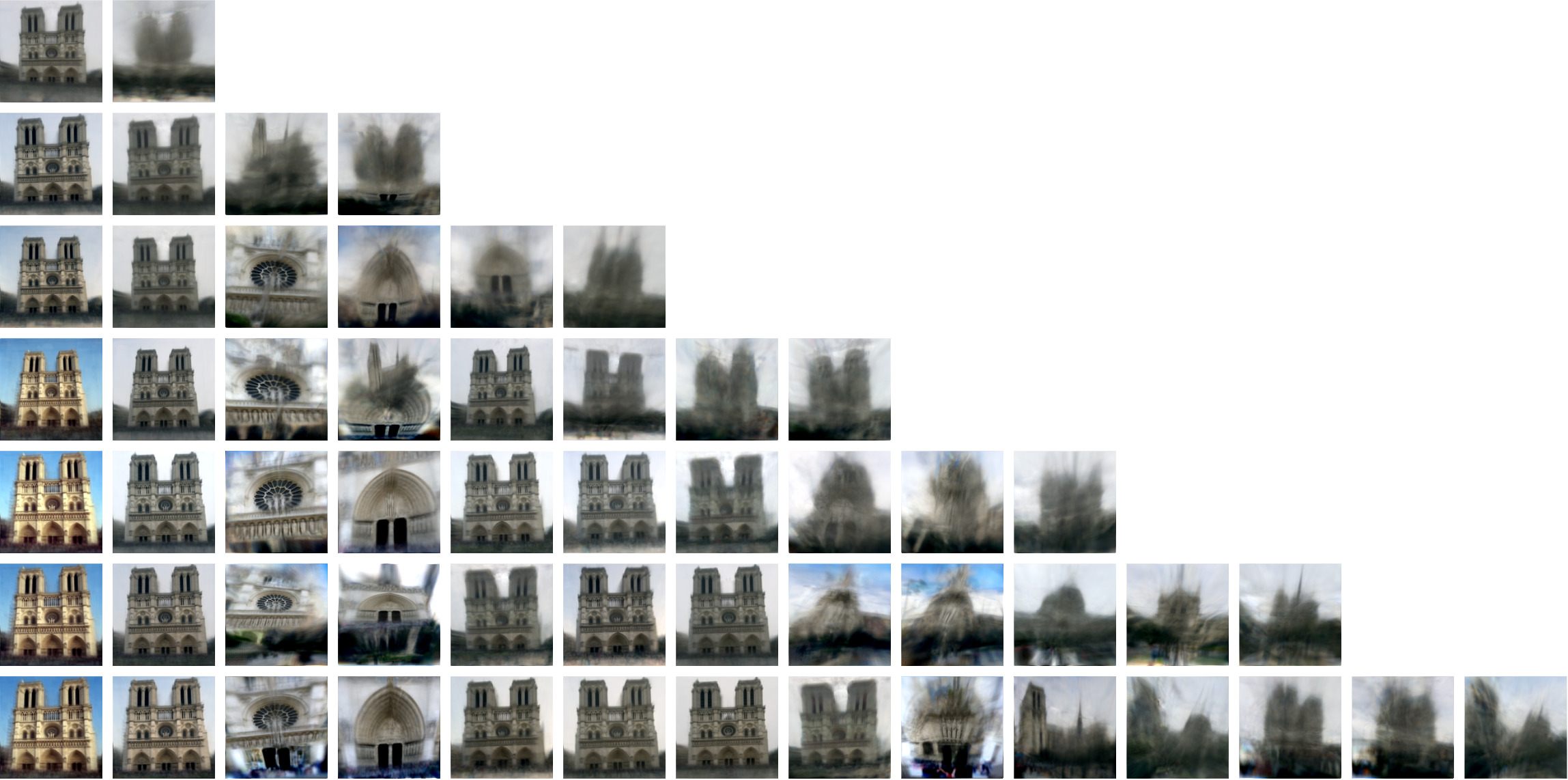}
    \caption{Prototypes learned on Notre-Dame for different numbers of clusters}
    \label{fig:loss_k_qual}
  \end{subfigure}

  \caption{\textbf{Effect of $K$. (a)} We report the loss of DTI K-means for different 
  numbers of clusters. For MNIST-test (left), the loss is averaged over 5 runs and for 
Notre-Dame (right), the loss corresponds to a single run. \textbf{(b)} We show the prototypes 
learned on Notre-Dame for even numbers of clusters.}
  \label{fig:loss_wrt_k}
\end{figure}

Similar to many clustering methods, the selection of the number of clusters is a challenge.  
We investigated if a purely quantitative analysis could be applied to select $K$. In 
Figure~\ref{fig:loss_k_quant}, we plot the loss of DTI-Kmeans as a function of the number of 
clusters for MNIST-test (left) and Notre-Dame (right). For MNIST-test, it is clear an elbow 
method could be applied to select the good number of clusters. For Notre-Dame, the 
quantitative analysis is not as conclusive but in this case, the correct number of clusters 
is not clearly defined. In practice, we did not find the qualitative results on internet 
photo collections to be very sensitive to this choice, as shown in 
Figure~\ref{fig:loss_k_qual} where learned prototypes are mostly consistent across the 
different clustering results.

\subsection{Constraining color transformation}\label{sec:color_discuss}

\begin{figure}[!h]
  \centering
  \includegraphics[width=0.55\columnwidth]{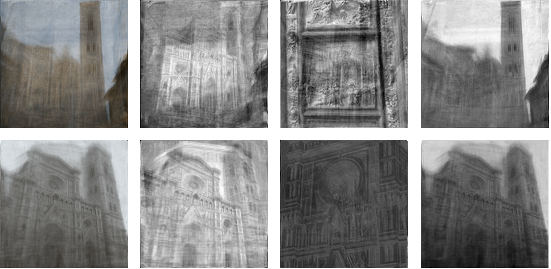}
  \caption{\textbf{Learned prototypes and RGB decomposition.} Two examples of learned 
    prototypes (first column) on Florence cathedral collection from~\cite{li2018megadepth} 
  using a full color transformation module (12 parameters). The 3 right columns correspond to 
R, G and B channels rescaled between 0 and 1. Note how the green channel is used to hide a 
completely different pattern from the other 2 channels.}
  \label{fig:color_tsf}
  \vspace{-2em}
\end{figure}

While evaluating our approaches on real photograph collections, we experimentally observed 
that a full affine color transformation module (12 parameters) was too flexible and as a 
result, prototypes were able to learn different patterns hidden in each color channel. In 
Figure~\ref{fig:color_tsf}, we show each R, G and B channel as a greyscale image for two 
prototypes learned using a full affine color transformation module. One can see that a second 
pattern is hidden in particular in the green channels. To avoid this effect, we restricted 
the color transformation module to be a diagonal affine transformation corresponding to 6 
parameters in total.%

\end{document}